\documentclass[runningheads]{llncs}
\usepackage[T1]{fontenc}
\usepackage{graphicx,verbatim}
\usepackage{xspace}

\makeatletter
\DeclareRobustCommand\onedot{\futurelet\@let@token\@onedot}
\def\@onedot{\ifx\@let@token.\else.\null\fi\xspace}

\def\eg{\emph{e.g}\onedot} 
\def\ie{\emph{i.e}\onedot}

\def\etal{\emph{et al}\onedot}
\makeatother

\usepackage[noadjust]{cite} %

\usepackage{amsmath}

\usepackage{amssymb}
\usepackage{amsfonts}
\usepackage{caption}
\usepackage{subcaption}
\usepackage{hyperref}

\usepackage[capitalize]{cleveref}
\crefname{section}{Sec.}{Secs.}
\Crefname{section}{Section}{Sections}
\Crefname{table}{Table}{Tables}
\crefname{table}{Tab.}{Tabs.}

\usepackage{multirow}
\usepackage{makecell}
\usepackage{utfsym}
\newcommand{\crossmark}{\scalebox{0.75}{\usym{2613}}}

\usepackage{color}

\begin{document}

\title{U-Mamba2: Scaling State Space Models for Dental Anatomy Segmentation in CBCT}
\titlerunning{U-Mamba2: Multi-Anatomy CBCT Segmentation}
\author{Zhi Qin Tan\inst{1}\orcidID{0000-0002-5521-6808} \and
Xiatian Zhu\inst{2}\orcidID{0000-0002-9284-2955} \and\\
Owen Addison\inst{1}\orcidID{0000-0002-0981-687X} \and
Yunpeng Li\inst{1}\orcidID{0000-0003-4798-541X}}
\authorrunning{Z.Q. Tan et al.}
\institute{Centre for Oral, Clinical \& Translational Sciences, King's College London, United Kingdom\\
\email{\{zhi\_qin.tan,owen.addison,yunpeng.li\}@kcl.ac.uk} \and
Surrey Institute for People-Centred AI, University of Surrey, United Kingdom\\
\email{\{xiatian.zhu,yunpeng.li\}@surrey.ac.uk}}

\maketitle              %
\setcounter{footnote}{0} %

\begin{abstract}
Cone-Beam Computed Tomography (CBCT) is a widely used 3D imaging technique in dentistry, providing volumetric information about the anatomical structures of jaws and teeth. Accurate segmentation of these anatomies is critical for clinical applications such as diagnosis and surgical planning, but remains time-consuming and challenging.
In this paper, we present U-Mamba2, a neural network architecture designed for multi-anatomy CBCT segmentation in the context of the ToothFairy3 challenge. U-Mamba2 integrates the Mamba2 state space models into the U-Net architecture, enforcing stronger structural constraints for higher efficiency without compromising performance.
In addition, we integrate interactive click prompts with cross-attention blocks, pre-train U-Mamba2 using self-supervised learning, and incorporate dental domain knowledge into the model design to address key challenges of dental anatomy segmentation in CBCT.
Extensive experiments, including independent tests, demonstrate that U-Mamba2 is both effective and efficient, securing first place in both tasks of the Toothfairy3 challenge. 
In Task 1, U-Mamba2 achieved a mean Dice of 0.84, HD95 of 38.17 with the held-out test data, with an average inference time of 40.58s. In Task 2, U-Mamba2 achieved the mean Dice of 0.87 and HD95 of 2.15 with the held-out test data.
The code is publicly available at \url{https://github.com/zhiqin1998/UMamba2}.


\end{abstract}
    
\section{Introduction}
Cone-Beam Computed Tomography (CBCT) is a widely used imaging modality in dentistry. It provides comprehensive 3D volumetric information and excellent visualization of the orofacial region, including jaws, teeth, nerves \cite{cbct_review}. Accurate segmentation of individual anatomical structures in CBCT images is crucial in applications such as dental diagnosis, treatment, and surgical planning \cite{clinical_cbct_review,cbct_use_tyndall2012,cbct_endo}. However, manual segmentation of CBCT scans requires specialized domain expertise and is extremely time-consuming due to their three-dimensional nature \cite{toothfairy2}. Thus, there is a strong demand for robust and efficient CBCT segmentation algorithms to improve the accuracy and efficiency of dental care and ultimately lead to better patient outcomes.

Generally, network architectures for semantic segmentation can be categorized into three: 1) Convolutional neural networks (CNN) such as U-Net \cite{unet, nnunet} and DeepLab \cite{deeplab} with translation-invariant convolutions that can effectively capture hierarchical image features and are parameter-efficient with their shared kernel weights; 2) Transformers \cite{transformer} such as SETR \cite{setr} and SwinTransformer \cite{liu2021Swin} that treat images as a sequence of patches instead of extracting image features hierarchically to capture the global information better; and 3) Hybrid CNN-Transformer architectures such as nnFormer \cite{nnformer} and SwinUNETR \cite{swinunetr} that attempt to exploit the best of both worlds by combining their architectures.

While the hybrid architectures have improved the global feature capabilities of CNNs, transformers are highly resource-intensive due to the attention mechanism which scales quadratically with input size. This limitation reduces their suitability for healthcare applications, which often involve high-resolution 3D data and constrained computational resources in real-world settings. Recently, structured state space sequence models \cite{s4}, particularly the Mamba \cite{mamba} model, have emerged as an efficient and effective alternative to the transformer model. By selectively capturing relevant input features and scaling linearly with input size, Mamba outperforms transformers across multiple modalities \cite{mamba,s4nd,ssvm}. U-Mamba \cite{umamba} presented the first work to leverage Mamba for image segmentation, achieving superior performance and surpassing transformer-based networks in a range of medical image segmentation tasks. More recently, Dao \etal~\cite{mamba2} proposed Mamba2, based on the structured state-space duality (SSD) framework, which dramatically improves speed without weakening its performance.

In this paper, we propose U-Mamba2, a hybrid CNN-SSD architecture for 3D image segmentation. U-Mamba2 extends the previous U-Mamba model \cite{mamba} by leveraging the Mamba2 SSD framework that simplifies the Mamba architecture with stronger constraints imposed on the hidden space structure. Mamba2 introduced several architectural changes to enable tensor and sequence parallelism, providing a significant speedup without compromising performance. Similar to U-Mamba, U-Mamba2 can effectively extract local spatial features via CNN and capture global long-range dependencies with Mamba2. 
We implement interactive click prompts with cross-attention blocks and incorporate several domain knowledge to address key challenges of dental anatomy segmentation in CBCT.
Our extensive experiments demonstrate the superior performance of U-Mamba2 for CBCT segmentation, outperforming previous alternatives and achieving first place
for Tasks 1 and 2 of the ToothFairy3 challenge.

\section{Method}
\begin{figure}[tb]
    \centering
    \begin{subfigure}{0.66\textwidth}
        \centering \includegraphics[width=\linewidth]{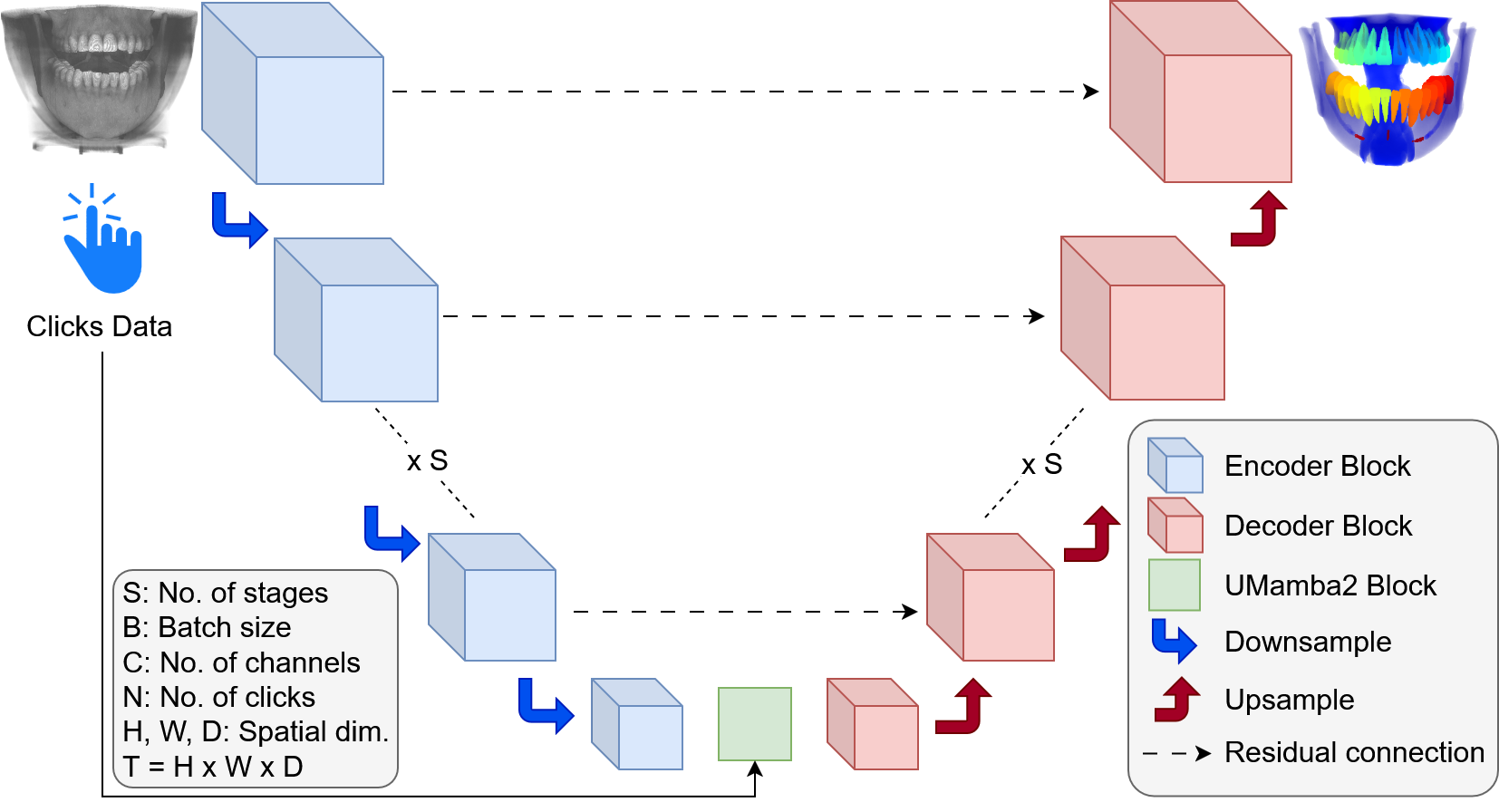}
    \end{subfigure} \hfill
    \begin{subfigure}{0.33\textwidth}
        \centering \includegraphics[width=\linewidth]{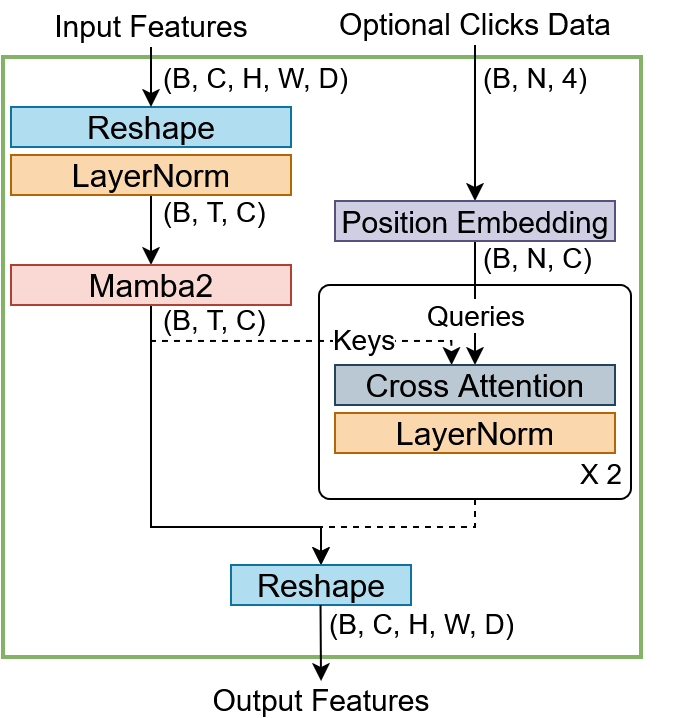}
    \end{subfigure}
    \caption{(Left): Overall architecture of the U-Mamba2 model. U-Mamba2 employs the encoder-decoder framework with residual connections between each stage and the U-Mamba2 block in the bottleneck. The number of stages is configurable depending on the dataset input size. (Right): The U-Mamba2 block contains the SSD-based Mamba2 and an optional click position encoder and cross-attention blocks. The output of Mamba2 follows the solid line for tasks without interactive clicks, while it follows the dashed line when clicks are present.}
    \label{fig:overall_arch}
\end{figure}

This section describes our method, designed in the scope of the two tasks of the ToothFairy3 \cite{toothfairy3,toothfairy3-ia,toothfairy3-tmi} challenges. Task 1 extends the previous ToothFairy2 challenge by adding segmentation of pulps, incisive nerves, and the lingual foramen to the existing 42 anatomy classes (\eg jaws, sinuses, and 32 teeth), and includes inference time to the evaluation criteria. The dataset contains 532 CBCT scans with shapes ranging from $(170, 272, 345)$ to $(298, 512, 512)$, along with the segmentation labels of 46 anatomy classes. On the other hand, Task 2 focuses on the interactive segmentation of the inferior alveolar nerves, allowing interactive user clicks as prompts to segment the inferior alveolar nerves. \cref{fig:overall_arch} shows the overall structure of the U-Mamba2 model and the details of the U-Mamba2 block.

\subsection{U-Mamba2: Integrating Mamba2 to U-Net}
Inspired by U-Mamba \cite{umamba}, we propose U-Mamba2, which integrates the strengths of U-Net and Mamba2 to efficiently capture global information. As shown in \cref{fig:overall_arch}, U-Mamba2 follows a structure similar to U-Net, with a symmetric encoder-decoder architecture that extracts image features across multiple scales. Residual connections between the encoder and decoder blocks at each stage facilitate the fusion of low-level and high-level features. As convolutional operations are inherently localized,
we leverage Mamba2 to enhance the vanilla U-Net's limited capability to model global long-range dependencies in images by treating the features as long sequences. Similar to Mamba, Mamba2 scales linearly with sequence length but leverages the SSD framework to constrain the internal recurrent structure and uses matrix multiplication instead of selective scan, thereby improving efficiency through parallelism.

The encoder blocks consist of two consecutive Residual blocks \cite{resnet}, followed by a strided downsampling convolution, while the decoder blocks are composed of Residual blocks and transposed convolutions for upsampling. In the U-Mamba2 block, image features of shape $(B,C,H,W,D)$ are reshaped and transposed to $(B,T,C)$ where $B$ denotes the batch size, $C$ the number of channels, and $H,W,D$ are the spatial dimensions, with $T=H\times W\times D$. Then, Layer Normalization \cite{layernorm} is applied to the features before they are passed to Mamba2 to capture the global contexts. The output features are then reshaped and transposed back to $(B,C,H,W,D)$. We apply the U-Mamba2 block exclusively in the bottleneck stage, as it results in the best empirical performance for 3D computed tomography modality, consistent with Ma \etal~\cite{umamba}. Finally, Softmax is applied to the final decoder feature to produce the segmentation class probabilities, and U-Mamba2 is trained with a combination of cross-entropy loss and Dice loss.

\subsection{Cross-Attention with Point Encoder}
We introduce an optional interactive branch to enable the model to incorporate user-provided clicks to refine the output of U-Mamba2, improve accuracy, and support human-in-the-loop collaboration. Following the SAM2 framework \cite{sam2}, this branch employs a position embedding and two cross-attention blocks, as illustrated in \cref{fig:overall_arch}. The optional clicks data contain a varying number of $N$ clicks consisting of the X, Y, Z coordinates, and class labels. These clicks are first encoded with a learnable position embedding depending on their spatial positions and class labels. Next, the embedded click prompts and the output features of Mamba2 are fused through two-way cross-attention blocks as queries and keys, respectively. The cross-attention blocks, followed by Layer Normalization, are repeated twice to allow the model to integrate click information with the image features. The final output of the cross-attention block is then reshaped and transposed back to the original spatial dimensions.

\subsection{Pre-training with Self-Supervised Learning}
\label{subsec:pretrain}
Recent studies \cite{dae,Tang2021SelfSupervisedPO} have shown that pre-training models on large datasets with self-supervised learning (SSL) produces stronger models that can extract meaningful feature representations, leading to improved performance of downstream segmentation tasks, particularly when there is limited labeled data. 

In addition to the 532 scans of ToothFairy3, we utilize the STS-3D-Tooth \cite{sdtooth} dataset consisting of 371 unlabeled CBCT scans to pre-train U-Mamba2 with the disruptive autoencoder (DAE) \cite{dae} framework. DAE aims to reconstruct the original 3D volume after it is corrupted by several low-level perturbations. Specifically, we corrupt the input volume by randomly applying local masks, downsampling, and adding Gaussian noise to the input. The disrupted input is then passed through the U-Mamba2 to learn to reconstruct the original image with an L1 loss function. The pre-trained weights are then used to initialize U-Mamba2 (except for the weights of the optional interactive branch and the final segmentation layer) for effective downstream training.

\subsection{Domain Knowledge for Dental Anatomy Segmentation}
\label{subsec:domain_know}
\paragraph{Label Smoothing of Related Anatomies.}
Anatomies in the orofacial region are not always distinct and often share similar shapes and properties. For instance, similar tooth types (incisor, canine, molar, premolar) between left-right counterparts, as well as the inferior alveolar and incisive nerves, exhibit close structural relationships. Therefore, to guide the model in recognizing similar classes and their spatial relationship, we introduce label smoothing for related anatomies instead of learning with hard one-hot labels. For each pixel with class $k$, we set the $k$-th class's target probability to $0.9$ and distribute the remaining $0.1$ evenly across the related classes. Specifically, for each voxel with a ground truth class label, $k$, and a set of related classes, $S_{r}$, we first initialize a zero vector, $p$, as the soft label, then set $p_k=0.9$ and $p_r = \frac{0.1}{|S_r|}, \forall r \in S_r$. We apply this strategy to all anatomies with left-right counterparts, neighboring teeth, and to the inferior alveolar and incisive nerves.

\paragraph{Weighted Loss for Tiny Structures.}
ToothFairy3 introduced three additional classes corresponding to the left and right incisive nerves and the lingual foramen, which house thin, sensitive nerves in the mandible. These structures are considerably smaller than other anatomies in the dataset. We account for the volume differences by applying a class weight of $10$ to these three tiny classes, so that their contribution to the overall loss is not overshadowed by larger anatomies.

\paragraph{Left-Right Mirroring Augmentation.}
The findings of the previous ToothFairy2 challenge \cite{toothfairy2,nnunet_tf2} showed that left-right mirroring augmentation can degrade the model’s capability to reliably differentiate the left/right orientation. In dentistry, even dentists may struggle to identify a horizontally-flipped 2D image reliably without visual cues \cite{xray_leftright}, due to the structural symmetry between left/right anatomies in the sagittal plane. However, we can exploit this anatomical symmetry with careful pre-processing and post-processing, enabling left-right mirroring augmentation without reducing model performance. We propose to swap the class labels of anatomies opposite to the sagittal plane whenever left-right mirroring occurs during data augmentation (\eg `Upper left canine' and `Upper right canine'). Additionally, we also switch the predicted logits of the corresponding left/right anatomies if the image is mirrored in the left-right axis during test-time augmentation (TTA). With proper processing during training and inference, the number of possible axes combinations for mirroring augmentation is expanded from 3 to 7, substantially increasing the generalization capabilities and performance of U-Mamba2.

\paragraph{Post-processing.}
We incorporate anatomical priors of the orofacial region that voxels belonging to the same anatomy should be connected and not separated into blobs, as a post-processing step. Unlike the first-place solution \cite{nnunet_tf2} of the previous ToothFairy2 challenge, we perform post-processing to remove small predictions that are likely false positives based on the volume of the computed connected components \cite{cc3d} instead of the total volume of each class. Specifically, we select the threshold as the 0.5th percentile of the connected components' volume computed using the ground truth for each class. Importantly, this threshold is determined through the statistics of the ground truth rather than model predictions, ensuring that it is not model-specific. The threshold for each class is pre-computed using the entire ToothFairy3 training dataset.

\section{Experiment Results}
We implement U-Mamba2 with the nnU-Net \cite{nnunet} framework. We perform a 9:1 stratified train-validation split on the ToothFairy3 dataset to ensure the same proportion of data sources (with different fields of view and imaging machines) in the train and validation datasets. All models are pre-trained with SSL (\Cref{subsec:pretrain}) following the original training configuration of DAE \cite{dae}. Each model employs seven encoder-decoder stages, an input patch size of 128x256x256, the native voxel spacing of 0.3mm\textsuperscript{3} (leading to no downsampling or upsampling during model training and inference), and a batch size of 1. During training, we disable left/right mirroring augmentation for all models except U-Mamba2, while during inference, we use sliding window inference with a tile size of 0.5 and disable left/right mirroring in TTA for all models, including U-Mamba2, for fair comparison. Other hyperparameters follow the default values of nnU-Net. Model training and time computation are performed on an RTX4090 GPU. We evaluate the models with the Dice coefficient, the Hausdorff Distance at the 95th percentile (HD95), and the average inference time in seconds, where lower is better for all metrics except Dice.

\subsection{Quantitative Results}
\begin{table}[tb]
    \caption{Validation set evaluation metrics. $\dag$ indicates applying post-processing.}
    \label{tab:quantitative_results}
    \centering \setlength{\tabcolsep}{4pt}%
    \resizebox{\linewidth}{!}{
    \begin{tabular}{l|ccccc|ccccc}
         \hline
         \multirow{2}{*}{Model} & \multicolumn{5}{c|}{Task 1} & \multicolumn{5}{c}{Task 2} \\
         \cline{2-11}
          & Dice & HD95 & Dice$\dag$ & HD95$\dag$ & Time & Dice & HD95 & Dice$\dag$ & HD95$\dag$ & Time \\
         \hline
         SwinUNETR \cite{swinunetr} & 0.858 & 48.86 & 0.874 & 40.09 & 7.23 & - & - & - & - & -\\
         nnU-Net ResE \cite{nnunet} & 0.861 & 45.28 & 0.887 & 32.05 & \textbf{6.20} & 0.901 & 1.98 & 0.905 & 1.71 & \textbf{5.06} \\
         U-Mamba \cite{umamba} & 0.865 & 42.06 & 0.896 & 25.88 & 6.98 & 0.903 & 1.65 & \textbf{0.913} & 1.58 & 5.88 \\
         U-Mamba2 (ours) & \textbf{0.873} & \textbf{41.08} & \textbf{0.908} & \textbf{21.35} & 6.81 & \textbf{0.905} & \textbf{1.63} & \textbf{0.913} & \textbf{1.57} & 5.70 \\
         \hline
    \end{tabular}
    }
\end{table}

\Cref{tab:quantitative_results} compares the proposed U-Mamba2 with nnU-Net ResE \cite{nnunet}, U-Mamba \cite{umamba} which utilizes the original Mamba layer \cite{mamba}, and SwinUNETR \cite{swinunetr} on the ToothFairy3 dataset. For Task 2, we incorporate a point prompt encoder to nnU-Net ResE and U-Mamba at the bottleneck stage, similar to U-Mamba2. U-Mamba2 outperforms all benchmark models, achieving the best mean Dice score of 0.873 and 0.905 for Tasks 1 and 2, respectively. After applying post-processing, U-Mamba2 further improves to a mean Dice score of 0.908 and 0.913 for Tasks 1 and 2, respectively. U-Mamba2 delivers the best performance with an average inference time of 6.81 and 5.70 seconds per scan for the two tasks, demonstrating a slight speedup over U-Mamba.

\begin{table}[tb]
    \caption{Ablation Study of U-Mamba2 for the validation set of Task 1. ILN indicates the metrics for the left and right incisive nerves and the lingual nerve.}
    \label{tab:umamba2_abl}
    \centering \setlength{\tabcolsep}{5pt}%
    \resizebox{0.85\linewidth}{!}{
    \begin{tabular}{ccc|cccc}
         \hline
         \makecell{Label\\ Smoothing} & \makecell{Weighted\\ Loss} & \makecell{L/R\\ Mirroring} & Dice & HD95 & Dice (ILN) & HD95 (ILN) \\
         \hline
         \crossmark & \crossmark & \crossmark & 0.867 & 42.36 & 0.617 & 38.41 \\
         \checkmark & \crossmark & \crossmark & 0.872 & 40.74 & 0.628 & 38.15 \\
         \crossmark & \checkmark & \crossmark & 0.870 & 41.31 & 0.635 & 37.99 \\
         \crossmark & \crossmark & \checkmark & 0.871 & 41.20 & 0.642 & 36.48 \\
         \checkmark & \checkmark & \checkmark & \textbf{0.873} & \textbf{41.08} & \textbf{0.646} & \textbf{35.21} \\
         \hline
    \end{tabular}
    }
\end{table}

Furthermore, we perform an ablation study on U-Mamba2 by individually applying the dental domain knowledge introduced in \Cref{subsec:domain_know}, excluding the post-processing step (See \Cref{tab:quantitative_results} for post-processing results). \Cref{tab:umamba2_abl} shows that these techniques lead to small performance improvements. In particular, the weighted loss and left/right mirroring techniques improve the mean Dice score on the three tiny structures, \ie the left and right incisive nerves and the lingual nerve (ILN) from 0.617 to 0.635 and 0.642, respectively. When all three techniques are applied, U-Mamba2 achieves the best performance, with a mean Dice score of 0.873 and 0.646 for all classes and the ILN classes, respectively.

\subsection{Qualitative Results}
\begin{figure}[tb]
    \begin{subfigure}{0.48\textwidth}
        \centering 
        \includegraphics[width=0.53\linewidth]{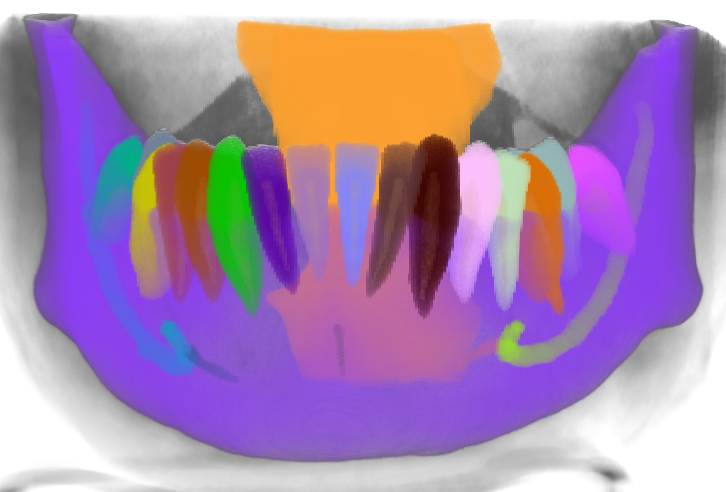}
        \includegraphics[width=0.45\linewidth]{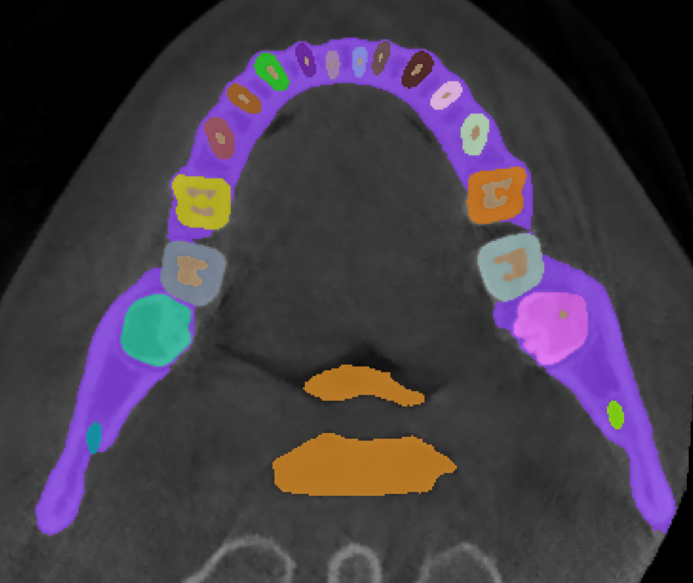}
    \end{subfigure} \hfill
    \begin{subfigure}{0.48\textwidth}
        \centering 
        \includegraphics[width=0.53\linewidth]{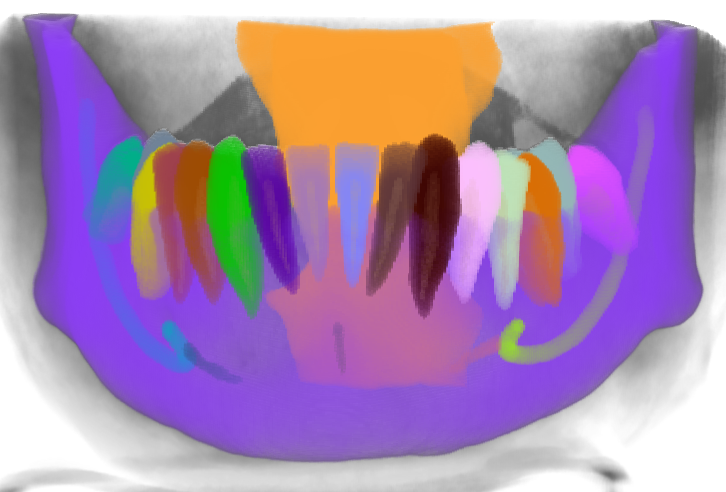}
        \includegraphics[width=0.45\linewidth]{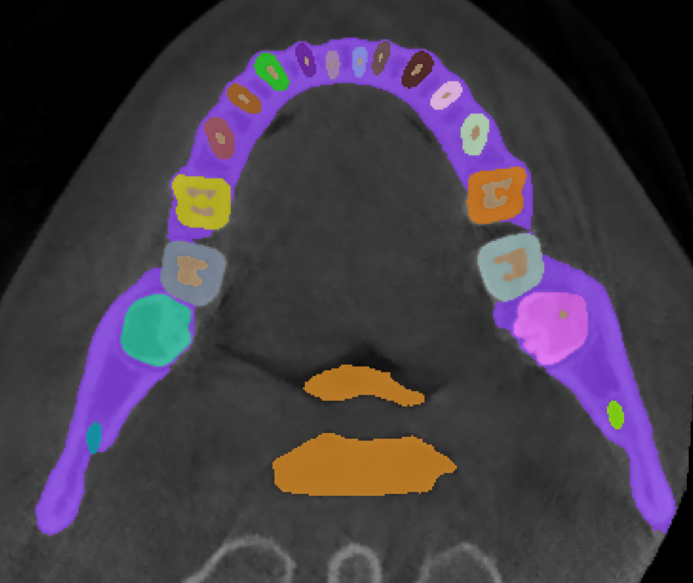}
    \end{subfigure}
    
    \begin{subfigure}{0.48\textwidth}
        \centering 
        \includegraphics[width=0.53\linewidth]{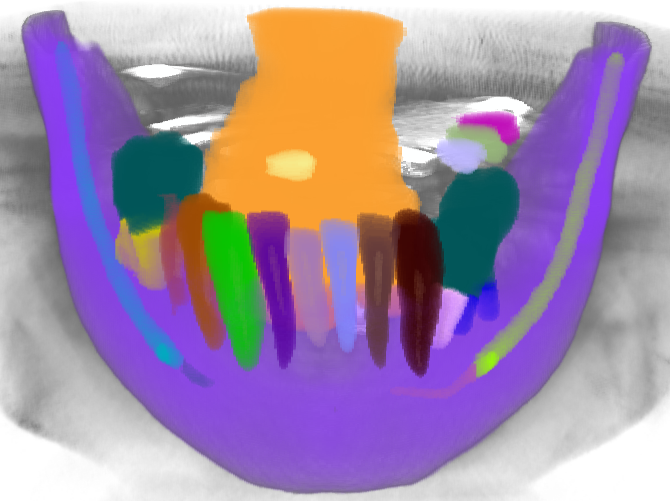}
        \includegraphics[width=0.45\linewidth]{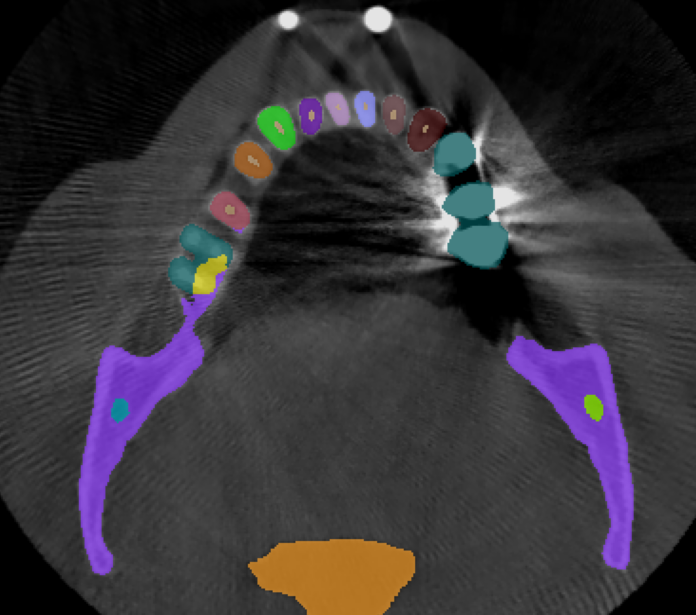}
        \caption{Ground Truth}
    \end{subfigure} \hfill
    \begin{subfigure}{0.48\textwidth}
        \centering 
        \includegraphics[width=0.53\linewidth]{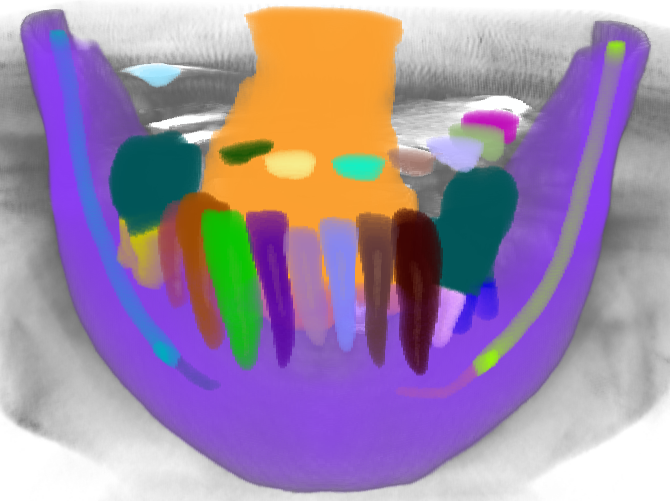}
        \includegraphics[width=0.45\linewidth]{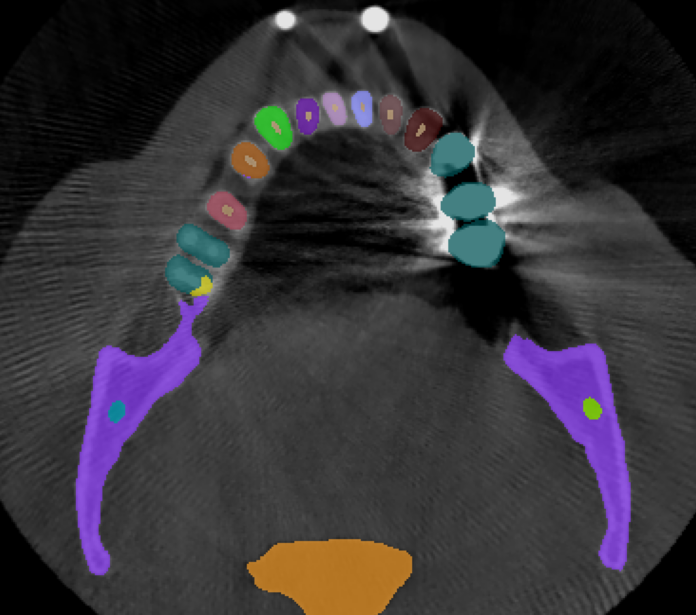}
        \caption{Prediction}
    \end{subfigure}
    \caption{Qualitative results of U-Mamba2 on the validation set of Task 1. The 3D render and a representative 2D slice are shown for: (Top) the best scoring case and (Bottom) the worst scoring case.}
    \label{fig:qualitative}
\end{figure}

\cref{fig:qualitative} visualizes the qualitative comparison between the ground truth and our model's predictions of the scans with the highest and lowest Dice score in the validation set, in the top and bottom rows, respectively. We observe that in most cases, U-Mamba2 produces precise segmentation predictions, showcasing the effectiveness of incorporating dental domain knowledge into the model design. Furthermore, we observe that U-Mamba2 can accurately localize the three tiny structures (ILN), producing visually acceptable segmentations. In the worst-case scenario, although the scan is imperfect due to image artifacts caused by metallic objects, false positives are primarily confined around the image edge or confusion between the actual tooth and the crown or implant, underscoring U-Mamba2's robustness under noisy conditions.

\subsection{Optimizing Speed in Sliding Window Inference}
\begin{figure}[tb]
    \begin{subfigure}{0.48\textwidth}
        \centering 
        \includegraphics[width=\linewidth]{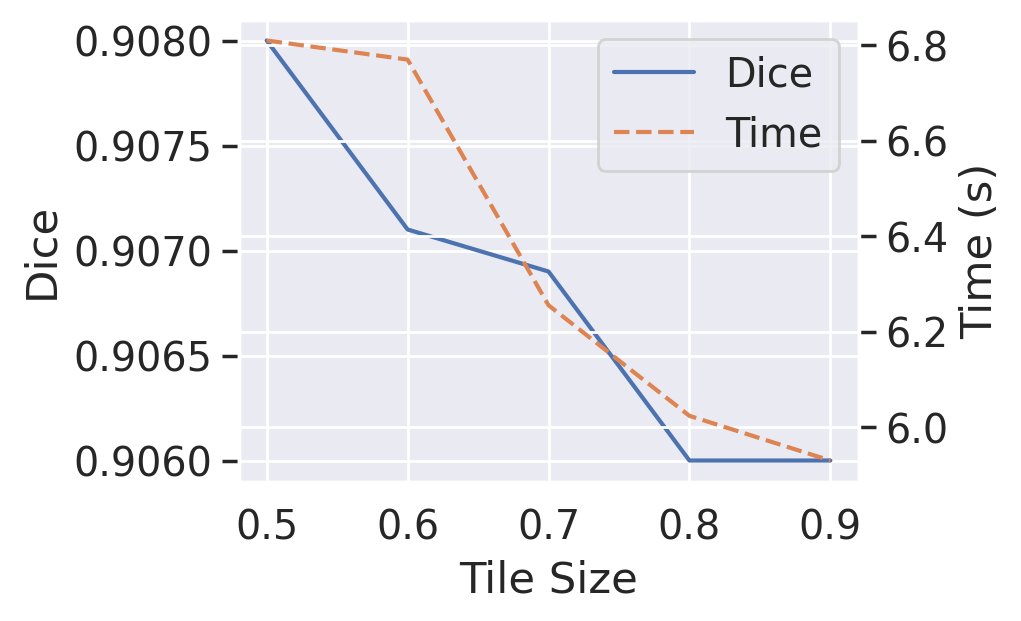}
    \end{subfigure} \hfill
    \begin{subfigure}{0.43\textwidth}
        \centering 
        \includegraphics[width=\linewidth]{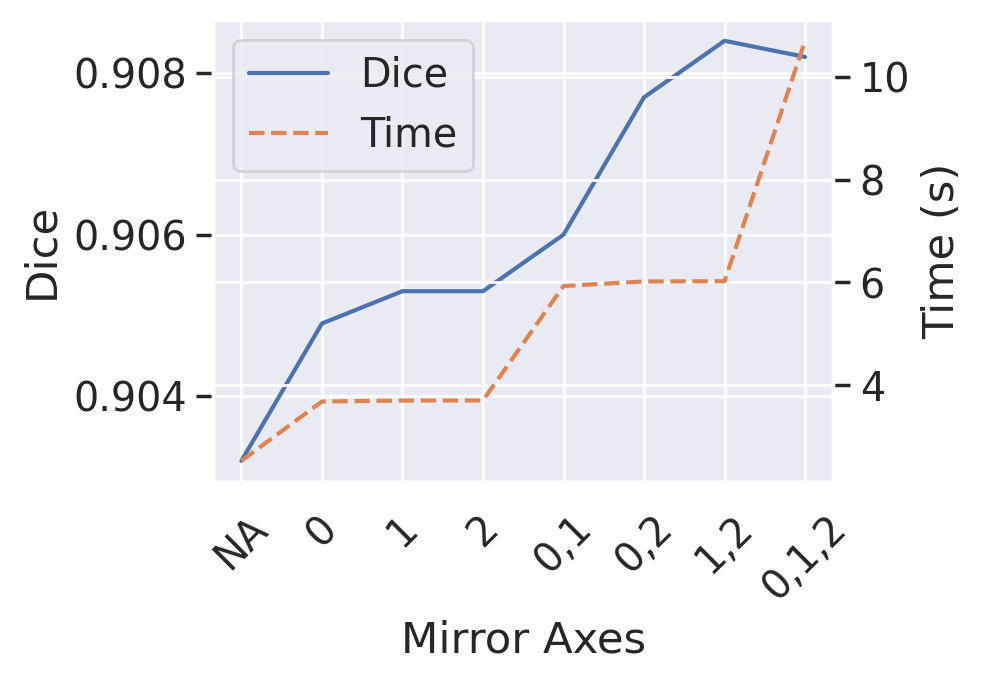}
    \end{subfigure}
    \caption{(Left): Effect of the tile size on the metrics with `0,1' mirror axes in TTA. (Right): Effect of various mirror axes combinations in TTA on the metrics when tile size is set to 0.9. Axis definition: `0' is superior/inferior, `1' is anterior/posterior, and `2' is left/right.}
    \label{fig:speed_tradeoff}
\end{figure}
As the inference time is an important metric in the ToothFairy 3 challenge, we optimize the sliding window inference parameters to improve speed without significantly deteriorating model accuracy. Specifically, we optimize the tile size parameter, where a larger value results in less border overlap during sliding window inference, and the mirror axes combinations in TTA. \cref{fig:speed_tradeoff} shows the tradeoff between Dice score and inference time for different tile sizes and mirror axes combinations in TTA. By setting the tile size to 0.9, we can reduce the inference time by 12.9\% with a negligible drop of only 0.002 Dice score. Moreover, \cref{fig:speed_tradeoff} also demonstrates that the optimal mirror axes combination is `1,2', representing anterior/posterior and left/right, offering the best Dice score with an average inference time of only 6.02 seconds. We believe this is due to the larger spatial dimension in these axes containing more information.

\subsection{Final Challenge Submission}
For the final submission, we extended training to 1500 epochs using all available data with a batch size of 2 and increased the input patch size to 160x288x288. During inference, we use a sliding window inference with a tile size of 0.9 and enable mirroring in the anterior/posterior and left/right axes during TTA.
The final U-Mamba2 model achieved a mean Dice of 0.84, HD95 of 38.17, with an average inference time of 40.58s, computed on the Grand Challenge platform using a T4 GPU, securing first place in Task 1 of the ToothFairy3 challenge with a 3.1 overall ranking while obtaining first place in Task 2 with a mean Dice, HD95 and overall rank of 0.87, 2.15 and 1.66, respectively, on the hidden test set.

\section{Conclusion}
We presented a new architecture, U-Mamba2, designed for multi-anatomy segmentation of CBCT images in the scope of the ToothFairy3 challenge. U-Mamba2 integrates the Mamba2 SSD framework into the U-Net backbone, achieving higher efficiency without compromising performance compared to U-Mamba. By incorporating domain-specific knowledge of dental anatomy, we improved the model's performance on multi-anatomy segmentation of CBCT scans. Both the validation and independent test results demonstrate the effectiveness and efficiency of U-Mamba2, securing first place in both Tasks 1 and 2 of the ToothFairy3 challenge.

\bibliographystyle{splncs04}
\bibliography{main}

\end{document}